\documentclass[conference]{IEEEtran}
\usepackage{amsmath,amsthm}
\usepackage{epsfig}
\usepackage{graphicx}
\usepackage{float}
\ifCLASSOPTIONcompsoc
  \usepackage[nocompress]{cite}
\else
  \usepackage{cite}
\fi
\usepackage[hyphens]{url}
\usepackage{color}
\usepackage[colorlinks=true,citecolor=black,urlcolor=black,linkcolor=black,hyperindex,breaklinks]{hyperref}
\usepackage[table,hyperref,x11names]{xcolor}
\usepackage{comment}
\usepackage{array}
\usepackage{hhline}
\usepackage[ruled,vlined,linesnumbered]{algorithm2e}
\usepackage{diagbox}
\usepackage{amssymb}
\usepackage{mathrsfs}
\usepackage{multirow}
\usepackage{threeparttable}
\usepackage{makecell}
\usepackage[font=small,skip=0pt]{caption}
\newcommand{\subparagraph}{}
\usepackage[hang]{footmisc}
\usepackage[normalem]{ulem}
\usepackage{scalerel}

\ifCLASSOPTIONcompsoc
    \usepackage[caption=false, font=normalsize, labelfont=sf, textfont=sf]{subfig}
\else
\usepackage[caption=false, font=footnotesize]{subfig}
\fi

\theoremstyle{definition}
\newtheorem{defn}{Definition}

\newtheorem{exmp}{Example}

\theoremstyle{remark}
\newtheorem{rem}{Remark}

\makeatletter
\def\hlinew#1{\noalign{\ifnum0=`}\fi\hrule \@height #1
\futurelet\reserved@a\@xhline}
\makeatother

\definecolor{greyf}{rgb}{0.7, 0.7, 0.7}
\definecolor{greys}{rgb}{0.85, 0.85, 0.85}

\newcolumntype{a}{>{\columncolor{Gray}}c}
\newcolumntype{b}{>{\columncolor{white}}c}

\newcommand{\etalc}[2]{{#1 \textit{et al.}\cite{#2}}}

\newcommand{\PreserveBackslash}[1]{\let\temp=\\#1\let\\=\temp}
\newcolumntype{C}[1]{>{\PreserveBackslash\centering}p{#1}}
\newcolumntype{R}[1]{>{\PreserveBackslash\raggedleft}p{#1}}
\newcolumntype{L}[1]{>{\PreserveBackslash\raggedright}p{#1}}

\begin{document}
\title{A Temporal-Pattern Backdoor Attack to Deep Reinforcement Learning}

\author{
\IEEEauthorblockN{Yinbo Yu\IEEEauthorrefmark{1}\IEEEauthorrefmark{2},
Jiajia Liu\IEEEauthorrefmark{1},
Shouqing Li\IEEEauthorrefmark{3},
Kepu Huang\IEEEauthorrefmark{1}, Xudong Feng\IEEEauthorrefmark{1}}
\IEEEauthorblockA{\IEEEauthorrefmark{1}School of Cybersecurity, Northwestern Polytechnical University, Xi'an 710072, China}
\IEEEauthorblockA{\IEEEauthorrefmark{2}Research \& Development Institute of Northwestern Polytechnical University in Shenzhen, Shenzhen 518057, China}
\IEEEauthorblockA{\IEEEauthorrefmark{3}Academy of Industrial Internet, Shaanxi Branch of China United Network Communications Group Co., Ltd., Xi’an, China}
\IEEEauthorblockA{Email: \{yinboyu, liujiajia\}@nwpu.edu.cn, lisq45@chinaunicom.cn, \{huangkepu, dfnjkd\}@mail.nwpu.edu.cn}
\\[-4.5ex]
}
\maketitle

\begin{abstract}
Deep reinforcement learning (DRL) has made significant achievements in many real-world applications. But these real-world applications typically can only provide partial observations for making decisions due to occlusions and noisy sensors. However, partial state observability can be used to hide malicious behaviors for backdoors. In this paper, we explore the sequential nature of DRL and propose a novel temporal-pattern backdoor attack to DRL, whose trigger is a set of temporal constraints on a sequence of observations rather than a single observation, and effect can be kept in a controllable duration rather than in the instant. We validate our proposed backdoor attack to a typical job scheduling task in cloud computing.  Numerous experimental results show that our backdoor can achieve excellent effectiveness, stealthiness, and sustainability. Our backdoor's average clean data accuracy and attack success rate can reach 97.8\% and 97.5\%, respectively.
\end{abstract}

\begin{IEEEkeywords}
Backdoor attack, deep reinforcement learning, temporal feature
\end{IEEEkeywords}

\IEEEpeerreviewmaketitle

% Manually equalize the lengths of two columns on the last page of your paper;
\section{Introduction}

Deep reinforcement learning (DRL) embraces deep neural networks (DNN) to overcome the limitations of reinforcement learning techniques in their convergence speed and performance, and has made significant achievements in various fields, e.g., networking, computer vision, and speech recognition \cite{luong2019applications}. The key factors to DRL success are large amounts of training data and increasing computational power. But existing models require expensive hardware and long training time. Hence, to reduce the overhead, users tend to use third-party resources to conduct training processes or directly use pre-trained networks provided by third-parties. However, this brings a new security threat of backdoor attack (also called ``neural trojan'', or ``trojan attack'') to perform DRL \cite{li2020backdoor, kiourti2020trojdrl}.

A backdoored DNN model behaves normally on benign samples, but can produce malicious results once an attacker-specified trigger is presented in the input. Unlike supervised learning (e.g., image classification), DRL is required to address sequential decision-making problems according to immediate rewards instead of supervision on long-term reward. Backdoors on DRL are more challenging since the backdoored agent needs to disrupt the sequential decisions rather than isolated decision while maintaining good performance in absence of backdoor triggers \cite{kiourti2020trojdrl}. \etalc{Kiourti}{kiourti2020trojdrl} use specific timesteps as the backdoor trigger and manipulate the corresponding action and reward when the trigger is present.  \etalc{Ashcraft}{ashcraft2021poisoning} studied a DRL backdoor that uses the action trajectory presented in the image observation as the trigger. While these backdoor triggers are useful for DRL, but lack stealthiness and are visible for users.

The DRL agent learns its optimal policy through interaction with the environment. Most DRL methods assumes that the state of the environment is fully observable for the agent. However, in many real world cases, the agent can only glimpse a part of the system state due to occlusions and noisy sensors. A Partially Observable Markov Decision Process (POMDP) \cite{kaelbling1998planning} can better capture the dynamics of many real world environments, but finding an optimal policy for POMDP is notoriously difficult. Existing methods aggregate histories of observations over time by integrating a recurrent neural network (RNN) \cite{hausknecht2015deep} or a generative model \cite{igl2018deep} to find hidden states for optimal policy generation. POMDP and these DRL methods have been used as a powerful tool to effectively address various problems and challenges, e.g., cognitive radio \cite{yang2020partially}, edge computing \cite{zhan2020deep}, automatic driving \cite{mo2021safe}, etc. However, the partial observability can make better hackers \cite{sarraute2012pomdps}. We think that a backdoor trigger can be hidden in unobservable states, which is invisible and hard to be detected.

Based on the above finding, in this paper, we propose a novel temporal-pattern trigger and a backdoor with controllable duration to DRL. A temporal-pattern trigger is a set of temporal constraints on a sequence of observations ($d_1,\cdots, d_t$) rather than a single observation. For instance, a trigger ($d_1-d_2<20\wedge d_3-d_2>10$) describes the correlation among three consecutive data. Such a trigger is invisible in a single epoch. Besides, most existing backdoor effects are activated instantly only when a trigger is present. \etalc{Yang}{yang2019design} presented a backdoor attack that persistently affects DRL performance once a trigger occurs. Persistent attack effect, however, destroys the stealthiness of the attack. We design a backdoor manipulating sequential decisions in a controllable duration. The adversary can specify the duration depending on the target DRL environment. With a temporal-pattern trigger and controllable attack duration, our DRL backdoor can achieve better stealthiness and practicality than existing works. To the best of our knowledge, this is the first that explores backdoor attacks to DRL in the time domain both regarding to trigger and attack effect. To evaluate our backdoor attacks, we use the task of job scheduling in cloud computing as a study case. Our experimental results demonstrate that the average clean data accuracy (CDA) and attack success rate (ASR) of our backdoor attacks is 97.8\% and 97.5\%, respectively.

The remainder of the paper is organized as follows. We discuss the related work in $\S$II. $\S$III provides the detailed insights on the DRL techniques applied in this research. We present our proposed backdoor in $\S$IV. $\S$V describes the experimental design and the performance evaluation results, and provides analytical insights. Lastly, we conclude the paper and outline our future work in $\S$VI.

\section{Related work}
\label{sec:relatedwork}

Backdoors in DNN have been widely studied in the domain of image classification \cite{li2020backdoor,gao2020backdoor}, where a trigger can be an image patch, a physical accessory, facial characteristic, or invisible noise. There are 6 major methods to insert backdoors into DNN: code poisoning, outsourcing, pretrained, data collection, collaborative learning, and post-deployment \cite{gao2020backdoor}.
While these experiences can be applied to DRL, DRL aims to solve sequential decision-making problems which are different from classification tasks.

Currently, there are only few works that have studied backdoors in DRL. \etalc{Kiourti}{kiourti2020trojdrl} use a timestep as the trigger and train a DRL policy to output fixed action and reward when the trigger is present. \etalc{Wang}{wang2021stop} design two backdoor attacks on DRL-based autonomous vehicles (AV) controllers. They set a specific set of combinations of positions and speeds of vehicles in the observation as the trigger. When the trigger is present, the AV controller could generate malicious deceleration to cause a physical crash and traffic congestion. We call these backdoors in these two work instant backdoors since their triggers and actions do not consider sequential dependencies.

Since the state space of many real-world environments is partial observable, introducing RNN into DRL can capture sequential dependencies to improve the performance of DRL. \etalc{Yang}{yang2019design} examine the persistent effect of backdoors in proximal policy optimization (PPO) algorithm equipped with LSTM for POMDP problems. They use a timestep $t$ as the trigger and setup a normal and a trojan environment to feedback the reward before and after $t$ respectively to train the backdoored policy. Similarly, \etalc{Ashcraft}{ashcraft2021poisoning} also study backdoors in this network, but the trigger they used is a sequence of action following a specific pattern. These two backdoor attacks are only applicable to simple games with totally tractable environments. \etalc{Wang}{wang2021backdoorl} study backdoors in reinforcement Learning used for two-player competitive games. They train two different policies (normal and Trojan), and hard-encode them to generate the backdoored trajectories which can lead to fast fail the game when a specific series of trigger actions (\textit{i.e.}, trigger) appears. Then, they use imitation learning to train an LSTM-based policy to mimic the behavior of generated trajectories. We also propose a backdoor attack on the LSTM-based DRL policy, but we study temporal constraints and controllable attack durations to build backdoors. These features ensure the practicality and stealthiness of our attack to DRL.

Since existing DNN backdoor attacks are majorly designed for classification tasks, most existing backdoor defense mechanisms are geared towards classification networks \cite{li2020backdoor,gao2020backdoor}. \etalc{Tran}{tran2018spectral} studied how to filter poisoned samples from the training set and demonstrated that poisoned samples tend to often contain detectable traces in their feature covariance spectrum.
STRIP \cite{gao2019strip} can detect whether an input contains trojan trigger by adding strong perturbation to the input. Neural Cleanse \cite{wang2019neural} identifies backdoored DNN models through reversing engineers for each input, so that all inputs stamped with the pattern are classified to the same label. However, these defense mechanisms are designed for isolated inputs rather than sequential inputs and have been shown low accuracy of detecting backdoors in DRL \cite{kiourti2020trojdrl}. Recently, \etalc{Guo}{guo2022backdoor} propose a backdoor detection method in competitive DRL by training a separate policy with a reversed reward function given by the Trojan agent. More sophisticated defense methods for DRL are desired to develop.

\section{Background}

\subsection{Partially-Observable Markov Decision Process}

Formally, a POMDP can be described as a 7-tuple $(\mathcal{S},\mathcal{A},\mathcal{O},\mathcal{P},\Omega, \mathcal{R},\gamma)$, where $\mathcal{S},\mathcal{A},\mathcal{O}, \mathcal{R}$ are the state space, action space, observation space, and rewards. In general, at each epoch $t$, an agent observes a state $s_t \in \mathcal{S}$ from the environment and then selects an action $a_t \in\mathcal{A}$ to the environment. The environment will perform $a_t$ and return a reward $r_t\in\mathcal{R}(s_t,a_t)$ to the agent. After, the agent obtains next state $s_{t+1}$ following the transition function  $\mathcal{P}(s_{t+1}|s_t,a_t)$. However, due to partial observability, the agent actually can only observe partial state $o_t\in \mathcal{O}$. The observation is generated by the environment following the observation function $\Omega(o_{t+1}|s_{t+1},a_t)$. $\gamma\in [0,1]$ is the discount factor and specifies the importance between future rewards and the current reward, \textit{i.e.}, $\gamma=0$ represents a ``myopic'' agent only concerned with its immediate reward, while $\gamma=1$ denotes an agent striving for a long-term higher reward.

\subsection{Deep Recurrent Q-Network}

To solve sequential decision problems, DRL is envisaged with the goal of maximizing the expectation of the long-term rewards. DRQN \cite{hausknecht2015deep} is one of the most popular DRL algorithms to address POMDP. Hence, in this paper, we use DRQN as the target DRL method. Under a POMDP, DRQN aims to learn estimates for the optimal value of each action, which is the expected sum of long-term rewards of executing the action from a given state. Hence, the agent keeps a parameterized value function $Q(s,a;\theta)$ for each state-action pair, where $\theta$ is the parameters of the network. Besides, DRQN also keeps a target network $\hat{Q}$ with parameters $\theta'$, which is the same as the online network $\theta$ except that $\theta'$ is copied every $C$ steps from $\theta$. Based on the value function, the agent decides an action to maximize the discounted long-term cumulative rewards which can be described as follows  \cite{hausknecht2015deep}:
\vspace{-1.5mm}
\begin{equation}\label{equ:target}
  y_t=r_{t+1}+\gamma\max_{a_{t+1}}\hat{Q}(s_{t+1},a_{t+1};\theta').
\end{equation}
\noindent The Q-value function is learned iteratively by updating the current value estimate towards the returned reward and estimated utility of the resulting states by the following expression:
\vspace{-1.5mm}
\begin{equation}
  Q(s,a;\theta) = Q(s,a;\theta)+\alpha(y- Q(s,a;\theta)),
\end{equation}

\noindent where $\alpha$ is the learning rate.

Additionally, DRQN includes a LSTM layer into its multi-layers neuron networks. LSTM is the most effective recurrent neural network architecture for prediction and classification. It can summarize the history for DQN by remembering features of the past, so as to help DQN better approximate actual Q-values from observation sequences, leading to better policies for partial observable environments \cite{hausknecht2015deep}.

\section{Temporal-Pattern Backdoor Attack}

\subsection{Threat model}

\textbf{Attacker’s Capacities}. Similar to existing DRL backdoors \cite{yang2019design,kiourti2020trojdrl,ashcraft2021poisoning}, we consider the outsourcing scenario, in which a user defines the model architecture, provides training data, and outsources the model training to a third party due to her lack of DRL skills or computational resources. In this scenario, attackers are allowed to poison some training data, control the training process, and manipulate the reward to the poisoning state. The security threat exists in many DNN-based applications, including DRL, autonomous driving, and face recognition \cite{gao2020backdoor,gong2021defense}.

\textbf{Attacker’s Goals}. With these above capacities, backdoor attackers aim to embed hidden backdoors into DRL policies. In particular, these inserted backdoors should have effectiveness, stealthiness, and sustainability \cite{gao2020backdoor}. The effectiveness requires that the backdoored model can perform no different from a normally-trained model in absence of backdoor triggers and have degraded performance when the trigger is present; The stealthiness requires that backdoor triggers should be concealed and have a small poisoning rate; The sustainability requires that the attack should still be effective under some common backdoor defenses.

\subsection{Temporal-Pattern Trigger}

The DRL agent is typically driven by fluctuating states in many real-world cases, e.g., job requests of cloud resources \cite{wei2018drl}, network bandwidth for adaptive video streaming \cite{mao2017neural}, and surrounding vehicle information \cite{mo2021safe}. These states are not affected by DRL decisions and thus, are uncontrollable for the DRL agent. But these states may be possible to manipulate by malicious users or attackers. For example, a cloud user builds a series of service requests containing a backdoor trigger to send to the cloud provider who uses a DRL model for service provision. Hence, we aim to hide backdoor triggers into these states, which ensure that the backdoor is realistic and cannot be removed by adjusting actions generated by the DRL policy. We define our \textit{temporal-pattern trigger} as follows:

These fluctuating states are time series data and can be denoted by $s^{ts}_0, s^{ts}_1, \cdots, s^{ts}_t, \cdots, s^{ts}_{\mathbb{N}}$, where $t\in\mathbb{N}$ is the timestep at which the DRL agent makes a decision (\textit{i.e.}, action $a_t$) according to the state $s_t$ provided by the environment. The state $s_t$ includes the fluctuating state $d_t$ and other partial states of the system. We encode backdoor triggers in the form of temporal logic formula over a sequence of time series data. Given two time series states $s^{ts}_1$ and $s^{ts}_2$, we define a formula $\varphi_{(1,2)} := f(s^{ts}_1, s^{ts}_2)\sim c$ to describe a constraint over $s^{ts}_1$ and $s^{ts}_2$, where $f$ is a binary operation (e.g., $+, -, \times, \div$), $\sim\in\{\equiv,\neq, >, \geq, <, \leq\}$, and $c\in\mathbb{R}$ is constant. The syntax of a temporal-pattern trigger is defined as follows:

\begin{defn} Given a set ($s^{ts}_{t-\mathbb{N}_t+1}, s^{ts}_{t-\mathbb{N}_t+2}, \cdots, s^{ts}_t$) of time series states with length $\mathbb{N}_t$ ending at timestep $t$, a \textbf{temporal-pattern trigger} is a temporal logic formula $\tau$ over constraints $\Psi$ of these time series states:
\begin{equation}
  \tau := \varphi_{(i,j)}|\varphi_{(i,j)} \otimes \varphi_{(k,s)}|ite(\varphi_{(i,j)}, \varphi_{(k,s)}, \varphi_{(p,q)}),
\end{equation}
\end{defn}
\noindent where $\varphi_{(i,j)}, \varphi_{(k,s)}, \varphi_{(p,q)} \in\Psi$; $\otimes$ is a Boolean operator, including disjunction $\vee$ (``or'') and conjunction $\wedge$ (``and''); $ite$ denotes $\phi$-assignment, e.g., $\tau := ite(\varphi_1, \varphi_2, \varphi_3)$ means if $\varphi_1$ is true, $\tau := \varphi_2$; otherwise $\tau := \varphi_3$. With the formula, attackers can define a backdoor trigger over time series data.
\begin{exmp}
For instance, given a trigger ($\mathbb{N}_t=4$) $\tau := \varphi_{(1,2)}\wedge \varphi_{(2,3)}\wedge ite(\varphi_{(1,3)}, \varphi_{(2,4)}, \varphi_{(4,1)})$, where $\varphi_{(1,2)} := d_1-d_2<0.01$, $\varphi_{(2,3)} := d_2-d_3>0.4$, $\varphi_{(3,4)} := d_3-d_4<-0.1$, $\varphi_{(4,2)} := d_4-d_2>-0.35$, $\varphi_{(4,1)} := d_4-d_1<0.1$, the normalized time series data (0.21, 0.83, 0.42, 0.47, 0.05, 0.17, 0.8) contains a trigger at 0.17 (\textit{i.e.}, 0.42, 0.47, 0.05, 0.17).
\end{exmp}

\begin{rem}
 Triggers of previous backdoor attacks to DRL are temporal-specific \cite{kiourti2020trojdrl} or observable in a single observation \cite{yang2019design,ashcraft2021poisoning} rather than over a period of time.
\end{rem}

\subsection{The Proposed Attack}

In this section, we illustrate our proposed attack. We first describe how to generate temporal-pattern triggers and then present how to train the backdoored DRL model with the pseudocode shown in Alg. \ref{alg:tpb}.

\begin{algorithm}[!t]
\caption{Temporal-Pattern Backdoored DRQN.}\label{alg:tpb}
\linespread{0.95}\selectfont
\small
Initialize replay memory $D$\ to capacity $M$, temporal-pattern trigger $\tau$($\mathbb{N}_t$), backdoor duration $L$, poisoning rate $\lambda$\;
Initialize Q-Network $Q$, Target-Network $\hat{Q}=Q$ \;
\For{$episode =1 $ \text{to}$\ max\_training\_steps$}{

Randomly choose a set $\Sigma$ of $\mathbb{N}_t$ timesteps to inject random  triggers based on $\tau$, where $\frac{|\mathbb{N}_t||\Sigma|}{|\mathbb{N}|}<\lambda$\;
$poisonDur =0$\;
\For{$t=1$ \text{to} $\mathbb{N}$}{
  With probability $\epsilon$ select a random action $a_t$, otherwise select $a_t = \max_a Q(s_t,a)$\;
  Execute action $a_t$ and observe reward $r_t$, new state $s_{t+1}$, and terminal signal $d_t$\;
  \If{$t$ is the last timestep of a trigger in $\Sigma$}{
    $poisonDur = L$\;
  }
  \If{$poisonDur>0$}{
     $r_t = 1-r_t$\;
     $poisonDur = poisonDur -1$\;
  }
  Store $(s_t, a_t, r_t, d_t, s_{t+1})$ into $D$\;
  Sample a sequential batch $(s_i, a_i, r_i, d_i, s_{i+1})$ from $D$\;
  Calculate target $y_i$ according to Equ. (\ref{equ:target})\;
  Do a gradient descent step with loss $\|y_i-Q(s_i,a_i)\|^2$\;
  Every $C$ steps reset $\hat{Q}=Q$\;
}
}
\end{algorithm}
\setlength{\textfloatsep}{0pt}%

First, attackers specify a temporal-pattern trigger $\tau$ over a series of data with $\mathbb{N}_t$ timesteps (\textit{i.e.}, $\tau(\mathbb{N}_t)$) and a backdoor duration $L$. Note that although with LSTM, DRQN can have long memory effects, LSTM is not that ideal and require more complex neurons and longer training time for memorizing longer historical features \cite{zhao2020rnn}. Hence, we keep only $\mathbb{N}_t$ and $L$ no more than 10 timesteps. Solutions, like \cite{igl2018deep, zhao2020rnn}, can address this limitation and thereby, make our backdoor attacks have longer effects, but are out of our scope. Then, attackers randomly choose a set $\Sigma$ of $\mathbb{N}_t$ timesteps in the time series data with length $N$, in each of which the time series data is randomly changed to satisfy with $\tau$ (Line 4). To keep the balance between CDA and ASR of the backdoored model, we set a poisoning rate $\lambda$ ensure $\frac{|\mathbb{N}_t||\Sigma|}{|\mathbb{N}|}<\lambda$.

During each training iteration, in general, the agent selects an action at each timestep using the $\epsilon$-greedy method (Line 7), stores interactions (Line 8 and 14), and updates the network with a MSE loss $\|y_i-Q(s_i,a_i)\|^2$ (Line 14-18). Once the current timestep $t$ is the last timestep of a trigger in the generated $\Sigma$ set, we change the reward $r_t\in[0,1]$ returned from the environment to be $1-r_t$ within a poisoning duration $L$ (Line 9-13) as our backdoor behavior. This behavior leads the network to being updated in the opposite direction of the clean training process, thereby is undesired. Besides, at each time to update the network, we sample a sequential batch from the memory $D$ rather than a random batch. Such batch can help DRQN learn more sequential features and also ensure the ASR of our backdoor attacks.

\section{Evaluation}
\label{sec:experiment}

\subsection{Case study}
In our experiment, we use the typical job scheduling problem in clouds as the study case. The scheduling system consists of users, job queues, job scheduler, and virtual machines (VMs). A job (i.e., user request) can be represented by a 4-tuple $(J^{ID}, J^{AT}, J^{TP}, J^{SZ})$ describing its id, arrival time, type (I/O or computing intensive), and instruction length respectively. To process jobs, the system provides $\mathcal{V}$ VM instances, each can be denoted by a tuple $(V^{ID}, V^{TP}, V^{PS})$ representing its id, instance type, average instruction processing speed respectively. At each timestep $t$, the job scheduler takes a user job $J_i$ ($i\in\mathbb{I}$) and assign to a VM instance $V_j$. $J_i$ will enter a waiting FIFO queue of the instance. Each VM can only handle one job at a timestep. Hence, the response time ($J^{RT}_{ij}$) of $J_i$ is composed of \textit{waiting} time ($J^{WT}_{ij}$) in the queue and \textit{execution} time ($J^{ET}_{ij}$) of VM. To encourage the matching of $J^{TP}_i$ and $V^{TP}_j$, we design $J^{ET}_{ij}=J^{SZ}_i/(V^{PS}_j(V^{TP}_j\oplus J^{TP}_i+1))$, where $\oplus$ is an exclusive or operator and is used to punish mismatched job assignment (\textit{i.e.}, $J^{TP}_i\neq V^{TP}_j$). $J^{WT}_{ij}$ is calculated according to the status of the queue: if the queue is empty, $J^{WT}_{ij}=0$; otherwise, $J^{WT}_{ij}=V_{ij}^{AVA}-J_i^{AT}$, where $V_{ij}^{AVA}$ is the avaiable time of $V_j$ for job $J_i$. If $V_{i'j}^{AVA}$ of last job $J_{i'}$ in $V_j$ is greater than its $J_{i'}^{AT}$, $V_{ij}^{AVA} = J^{ET}_{i'j}+V_{i'j}^{AVA}$; otherwise, $V_{ij}^{AVA} = J^{ET}_{i'j}+J_{i'}^{AT}$.

%
% If $J_i$ is computing intensive, $J^{ET}_{ij}=J^{SZ}_i/V^{COM}_j$; otherwise $J^{ET}_{ij}=J^{SZ}_i/V^{IO}_j$.

To learn a DRL policy for addressing the above problem, we use the event-driven decision framework similar to \cite{wei2018drl, huang2021deep}. At each epoch $t$, the agent select an action for an incoming job $J_i$ from the action space $\mathcal{A} = \{a_i, i =1\cdots M\}$, \textit{i.e.}, which VM to assign $J_i$. The observation space is $\mathcal{O} = [J_i^{TP},J_i^{SZ},J_{i1}^{WT},\cdots,J_{i(M-1)}^{WT}]$, where $J_i^{TP}$ and $J_i^{SZ}$ are the current job $J_i$ ($i\in \mathbb{I}$) to be scheduled, and the remaining part are the state of waiting time for processing $J_i$ in all VMs. Note that we set the last $J_{iM}^{WT}$ to be unobservable. The scheduling system aims to minimize the response time of all jobs. Hence, we have the reward function $r=J_i^{SZ}/(J^{RT}_{ij}V^{PS}_i)$.

\subsection{Experiment Setup}

In our experiment, we set the number of VM instances to $\mathcal{V}=10$ and the number of jobs to $|\mathbb{I}|=1000$ (\textit{i.e.}, $\mathbb{N}=1000$). Similar to \cite{wei2018drl}, we have 5 computing intensive VM instances ($V^{TP}=0$) and 5 I/O intensive VM instances ($V^{TP}=1$) respectively, and the average processing speeds of these VM is 2000 MIPS. Besides, we set that job requests arrive according to a Poisson distribution and their arrival rate is varied from 10 to 30 requests/s. The instruction lengths of jobs obey the normal distribution with mean 200 MI and standard deviation 20 MI. The ratio of computing intensive jobs and I/O intensive jobs is 1:1. Our implementation is available at https://github.com/EboYu/DRLBackdoor.

We use the following linear DNN network: $\underbrace{|\mathcal{O}|\rightarrow 64}_{\text {Input}}\rightarrow\underbrace{64\rightarrow 64}_{\text {LSTM}\times 2}^{\curvearrowleft}\rightarrow \underbrace{64\rightarrow 64}_{\text {Hidden}}\rightarrow\underbrace{64\rightarrow |\mathcal{A}|}_{\text {Output}}$ to train a DRQN policy for job scheduling. The network contains a fully connected hidden layer with 64 neurons. Between the input and hidden layer, it has a LSTM network with 2 layers. Besides, we set the capacity of replay memory $M=10000$, the size of batch 64, discount factor $\gamma=0.9$, and steps of updating target network $C=50$. Adam algorithm is adopted to update the network $Q$ with the learning rate of $\alpha=0.001$. We set $\epsilon$ to be decreased from 0.9 by 0.002 in each learning iteration. We specify 4 temporal-pattern triggers to validate our backdoor attacks:
\begin{itemize}

  \item $\tau_1(\mathbb{N}_t=4, L=7):= \varphi_{(1,0)}\wedge \varphi_{(2,1)}\wedge\varphi_{(3,2)}\wedge\varphi_{(3,0)}$, where $\varphi_{(1,0)}:= d_1-d_0>-3 \wedge d_1-d_0<-2.6$, $\varphi_{(2,1)}:= d_2-d_1>90 \wedge d_2-d_1<100$, $\varphi_{(3,2)}:= d_3-d_2>-25 \wedge d_3-d_2<-12$, and $\varphi_{(3,0)}:= d_3-d_0>70 \wedge d_3-d_0<79$;

  \item $\tau_2(\mathbb{N}_t=4, L=4):= \varphi_{(1,0)}\wedge(\varphi_{(2,1)}^1\vee\varphi_{(2,1)}^2)\wedge\varphi_{(3,2)}$, where $\varphi_{(1,0)}:= d_1-d_0>80 \wedge d_1-d_0<81.8$, $\varphi_{(2,1)}^1:= d_2-d_1>8 \wedge d_2-d_1<10$, $\varphi_{(2,1)}^2:= d_2-d_1>-20 \wedge d_2-d_1<-17.2$, and $\varphi_{(3,2)}:= d_3-d_2>-50 \wedge d_3-d_2<-48.5$,;

  \item $\tau_3(\mathbb{N}_t=5, L=6):= \varphi_{(1,0)}\wedge\varphi_{(2,1)}^1\wedge\varphi_{(3,1)}
  \wedge ite(\varphi_{(2,1)}^2, \varphi_{(4,3)}^1, \varphi_{(4,3)}^2)$,
  where $\varphi_{(1,0)}:= d_1-d_0>20 \wedge d_1-d_0<20.8$, $\varphi_{(2,1)}^1:= d_2-d_1>-8 \wedge d_2-d_1<-3.5$,
   $\varphi_{(3,1)}:= d_3-d_1>-25 \wedge d_3-d_1<-22.5$, $\varphi_{(2,1)}^2:= d_2-d_1<-6$, $\varphi_{(4,3)}^1:= d_4-d_3>43 \wedge d_4-d_3<50$, and $\varphi_{(4,3)}^2:= d_4-d_3>-90 \wedge d_4-d_3<-85$.

   \item $\tau_4(\mathbb{N}_t=6, L=3):= \varphi_{(2,0)}\wedge\varphi_{(3,2)}\wedge\varphi_{(4,3)}\wedge\varphi_{(5,1)}$,
   where $\varphi_{(2,0)}:= d_2-d_0>89.5 \wedge d_2-d_0<90$, $\varphi_{(3,2)}:= d_3-d_2>-27 \wedge d_3-d_2<-26$,
    $\varphi_{(4,3)}:= d_4-d_3>5 \wedge d_4-d_3<9.3$, and $\varphi_{(5,1)}^2:= d_5-d_1<10 \wedge d_5-d_1>8$.
\end{itemize}

Our experiments are conducted on a machine with an Intel i9-10900K CPU and an Nvidia GTX 3090 GPU. We evaluate our backdoor attacks based on two standard metrics: the clean data accuracy (CDA) and attack success rate (ASR). Different from classification tasks, in our experiment, CDA ($R_{backdoored}/R_{normal}$) represents the gap between the performance $R_{backdoored}$ of the backdoored model and that $R_{normal}$ of a normally-trained model for solving the job scheduling problem. ASR ($N_{present}/N_{true}$) is the percentage that the performance of the backdoored model is degraded ($N_{present}$) when the trigger is present ($N_{true}$). Besides, since our backdoor has a controllable duration, we define a new metric, attack persistence rate (APR) ($1-|D_{designed}-D_{present}|/D_{designed}$), which is used to compare the number $D_{present}$ of timesteps that an attack can sustain after it appears with the designed duration $D_{designed}$.

\subsection{Numerous results}

\begin{figure}[t!]
\centering
\begin{minipage}[c]{1\columnwidth}
    \centering
    \subfloat[rate=10]{\includegraphics[width=0.5\columnwidth]{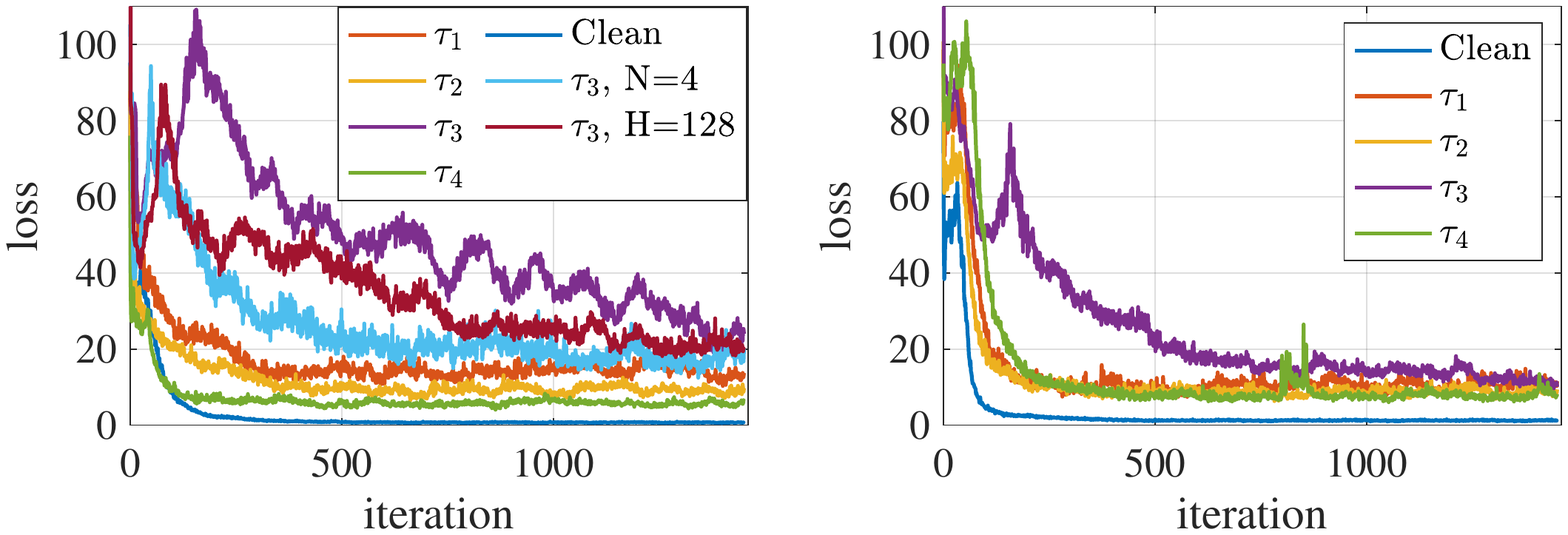}}
    \subfloat[rate=30]{\includegraphics[width=0.5\columnwidth]{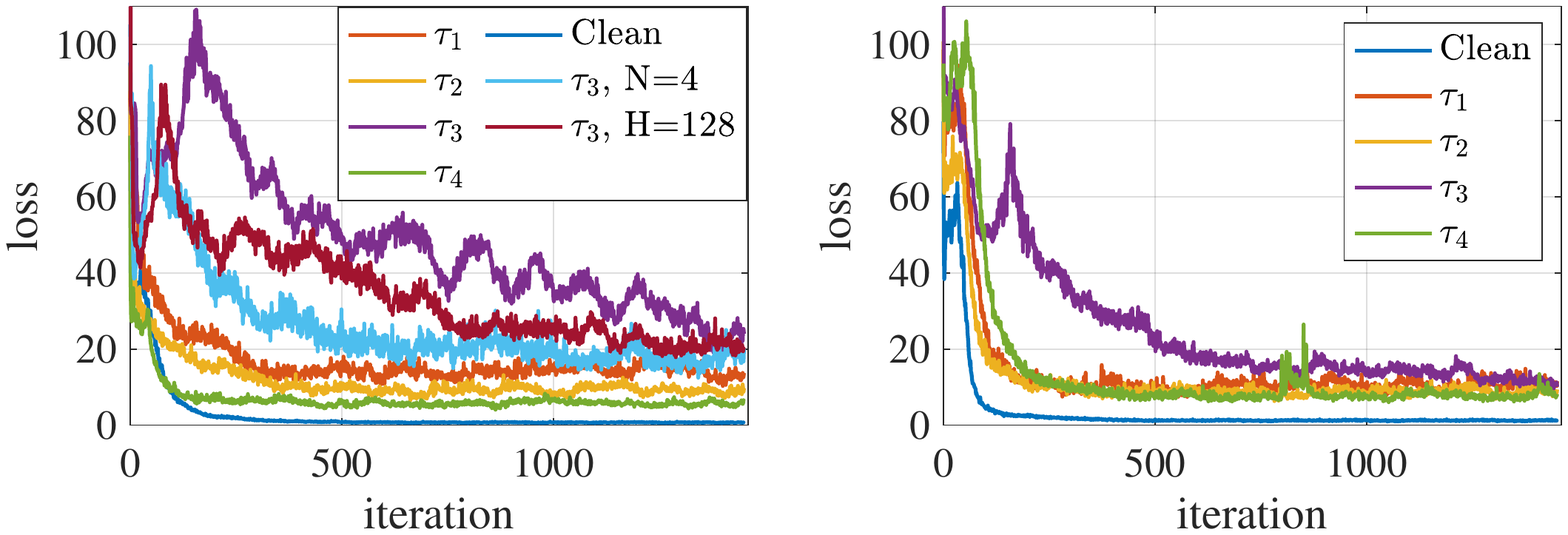}}
  \end{minipage}
  \caption{Convergence of training clean and backdoored models. H and N is the neuron size and number of LSTM layer, respectively.}
    \label{fig:conv}
\end{figure}

\begin{table}[t!]
  \caption{Average ASR/APR/CDA of four backdoor attacks}
  \label{tab:rate}
  \centering
  \begin{tabular}{c|cccc}
  \hline
  rate & $\tau_1$ & $\tau_2$& $\tau_3$& $\tau_4$\\\hline
  10 & 0.99/0.99/0.97& 0.97/0.97/0.98 & 0.91/0.93/0.96 & 0.99/0.99/0.99\\\hline
  15 & 0.98/0.98/0.97 & 0.99/0.98/0.98 & 0.97/0.97/0.97 & 0.99/0.99/0.98\\\hline
  20 & 0.99/0.98/0.96& 0.98/0.98/0.98 & 0.97/0.99/0.96 & 0.98/0.98/0.99\\\hline
  25 & 0.99/0.99/0.97& 0.97/0.96/0.98 &  0.96/0.97/0.97 & 0.99/0.98/0.98\\\hline
  30 & 1.0/1.0/0.97& 0.98/0.96/0.98 & 0.99/0.98/0.97 & 0.97/0.96/0.98\\\hline
  \end{tabular}
\end{table}

\begin{figure}[!t]
\centering
  \vspace{-4mm}
\begin{minipage}[c]{0.95\columnwidth}
    \centering
    \subfloat[$\tau_1(\mathbb{N}_t=4, L=7)$, rate=20]{\includegraphics[width=0.5\columnwidth]{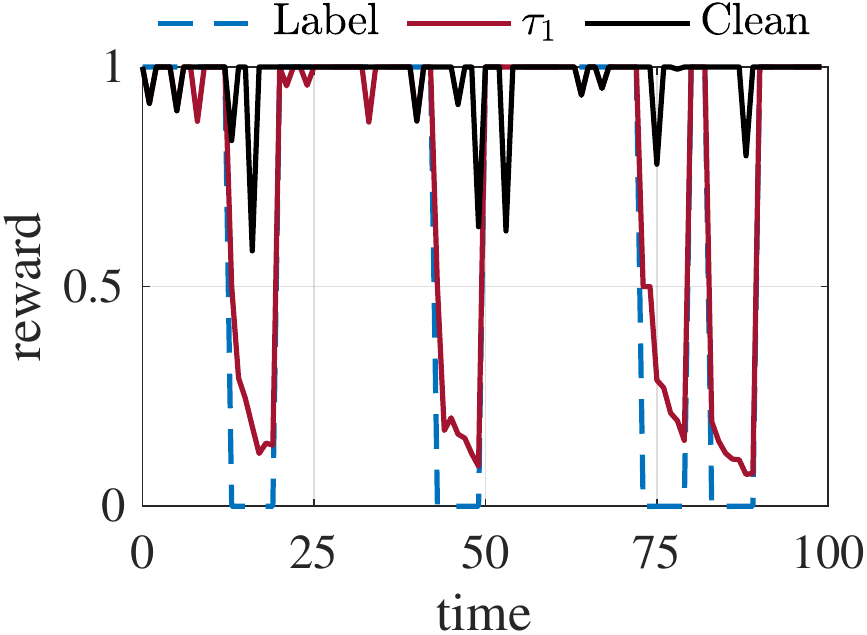}}
    \subfloat[$\tau_1(\mathbb{N}_t=4, L=7)$, rate=20]{\includegraphics[width=0.5\columnwidth]{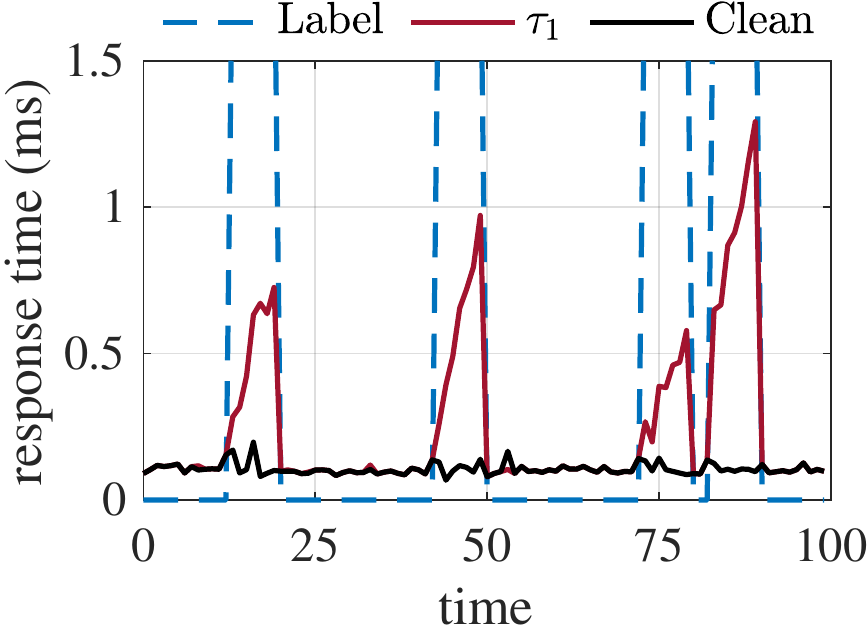}}
    \vspace{-3mm}
  \end{minipage}
  \vspace{-3mm}
  \begin{minipage}[c]{0.95\columnwidth}
      \centering
      \subfloat[$\tau_2(\mathbb{N}_t=4, L=4)$, rate=20]{\includegraphics[width=0.5\columnwidth]{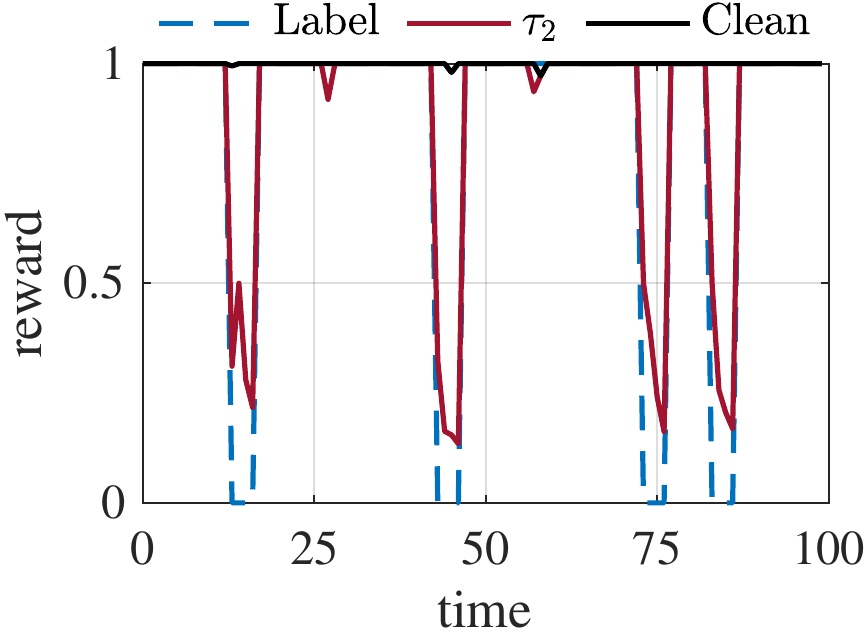}}
      \subfloat[$\tau_2(\mathbb{N}_t=4, L=4)$, rate=20]{\includegraphics[width=0.5\columnwidth]{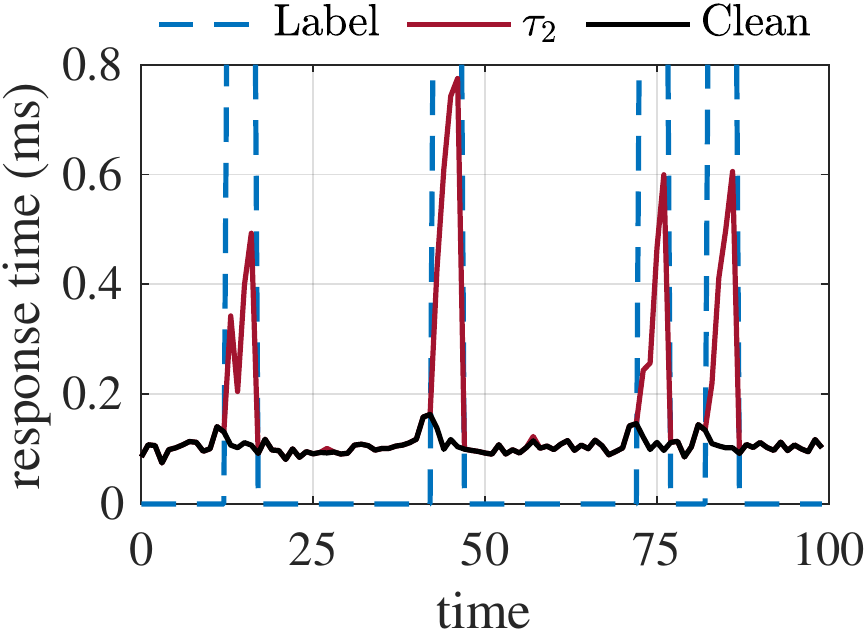}}
    \end{minipage}
\vspace{-3mm}
  \begin{minipage}[c]{0.95\columnwidth}
      \centering
      \subfloat[$\tau_3(\mathbb{N}_t=5, L=6)$, rate=20]{\includegraphics[width=0.49\columnwidth]{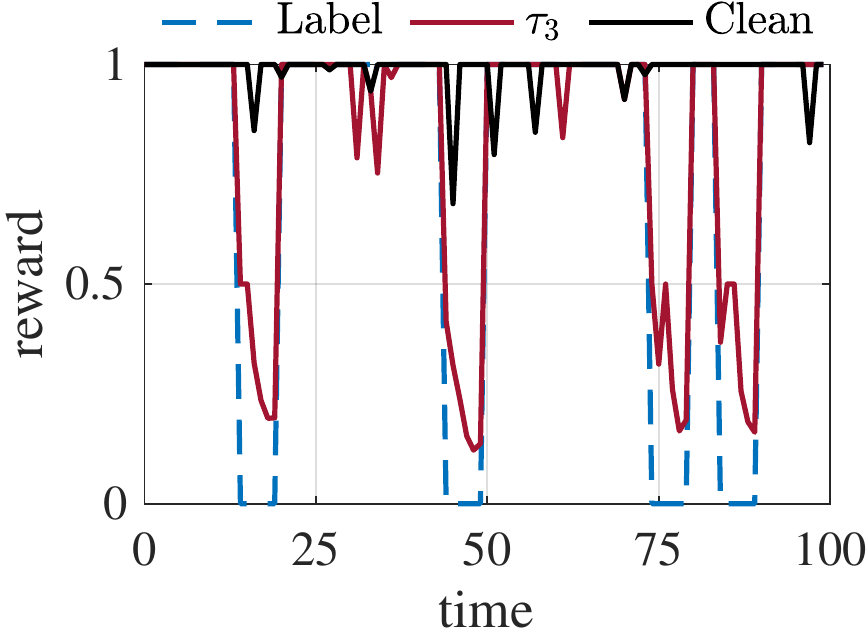}}
      \subfloat[$\tau_3(\mathbb{N}_t=5, L=6)$, rate=20]{\includegraphics[width=0.5\columnwidth]{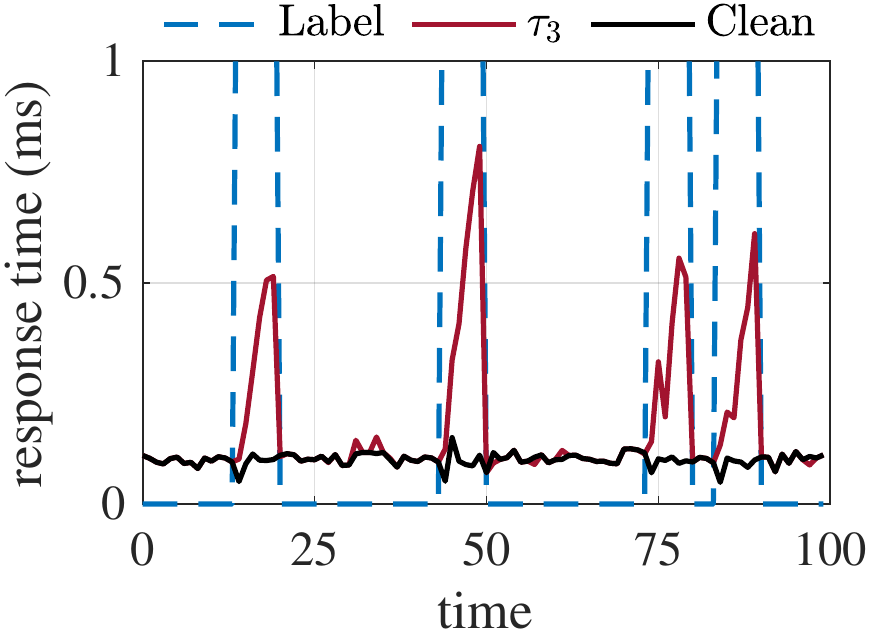}}
    \end{minipage}
\vspace{-3mm}
    \begin{minipage}[c]{0.95\columnwidth}
        \centering
        \subfloat[$\tau_4(\mathbb{N}_t=6, L=3)$, rate=20]{\includegraphics[width=0.5\columnwidth]{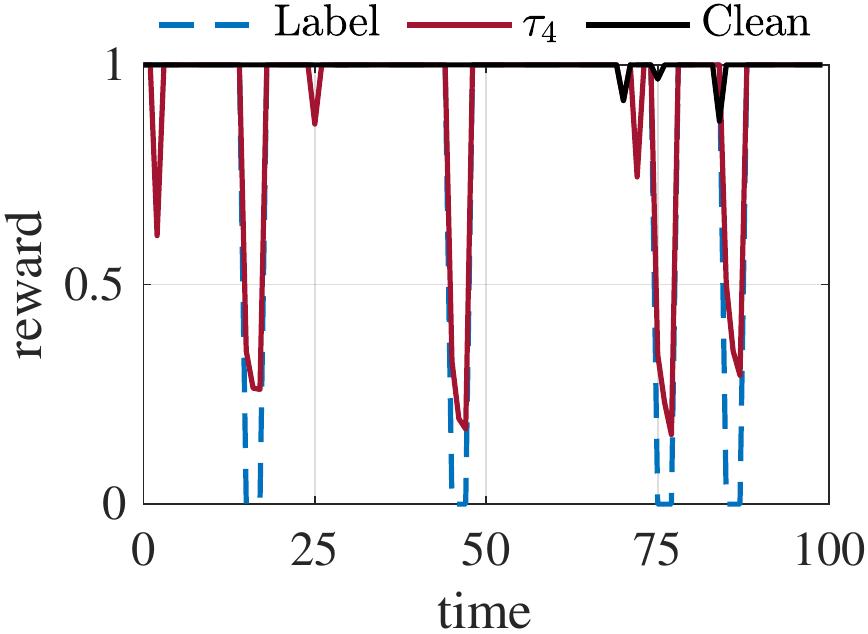}}
        \subfloat[$\tau_4(\mathbb{N}_t=6, L=3)$, rate=20]{\includegraphics[width=0.5\columnwidth]{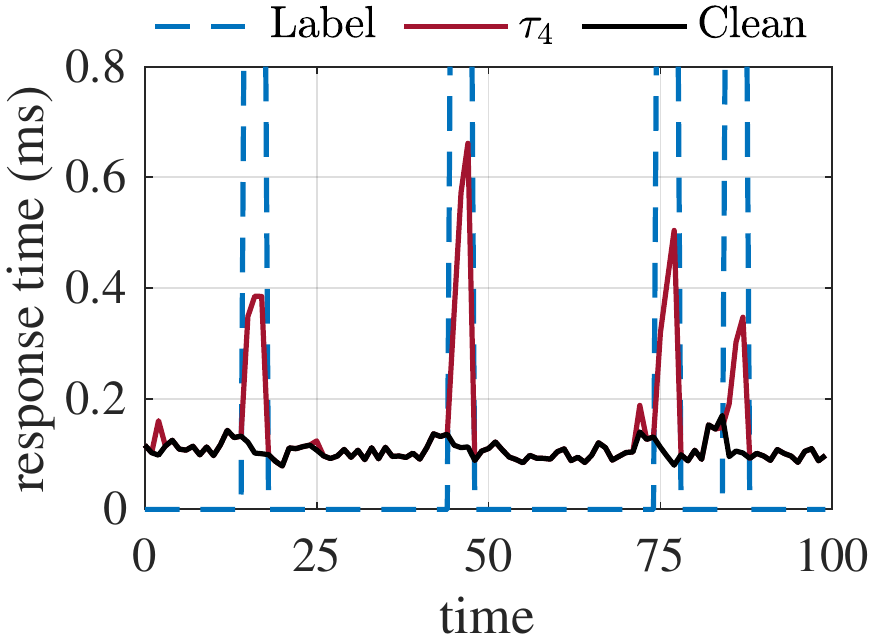}}
      \end{minipage}
\vspace{-3mm}
\begin{minipage}[c]{0.92\columnwidth}
    \centering
    \subfloat[$\tau_1(\mathbb{N}_t=4, L=42)$, rate=20]{\includegraphics[width=0.5\columnwidth]{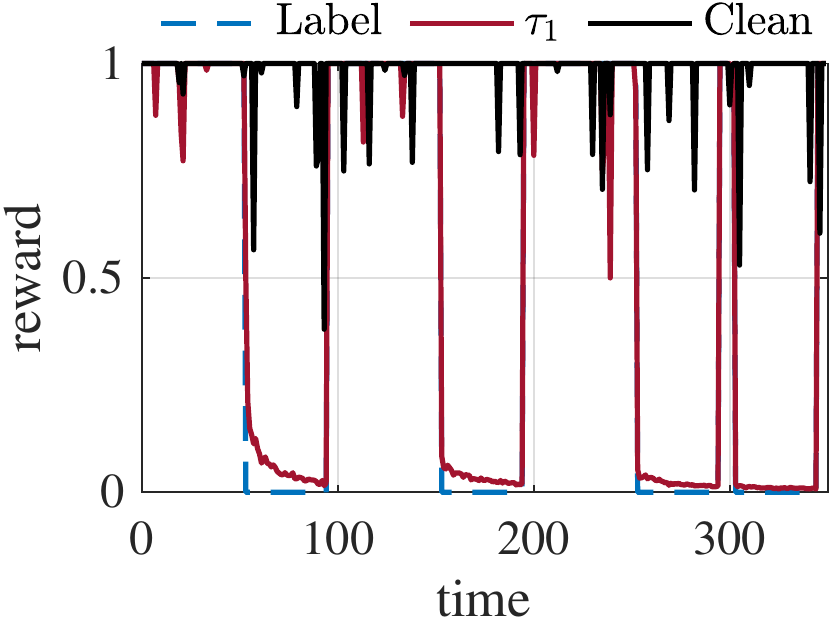}}
    \subfloat[$\tau_1(\mathbb{N}_t=4, L=42)$, rate=20]{\includegraphics[width=0.49\columnwidth]{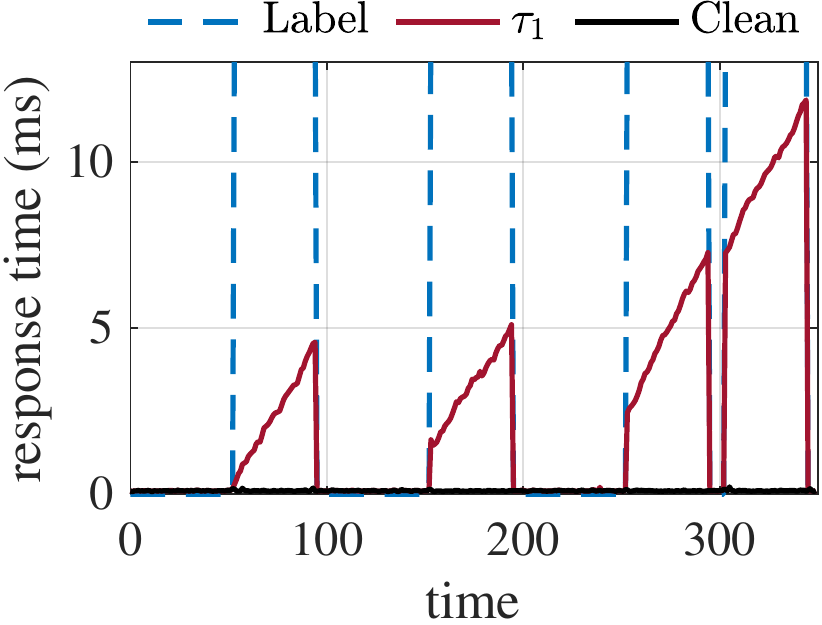}}
  \end{minipage}

\vspace{-0.8mm}
\begin{minipage}[c]{0.95\columnwidth}
    \centering
    \subfloat[$\tau_1(\mathbb{N}_t=4, L=7)$, rate=55]{\includegraphics[width=0.5\columnwidth]{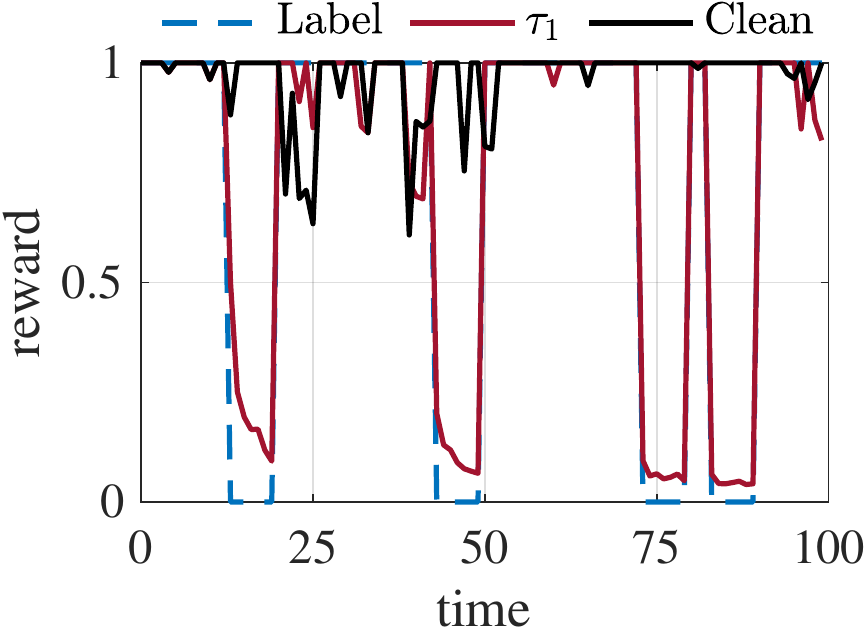}}
    \subfloat[$\tau_1(\mathbb{N}_t=4, L=7)$, rate=55]{\includegraphics[width=0.48\columnwidth]{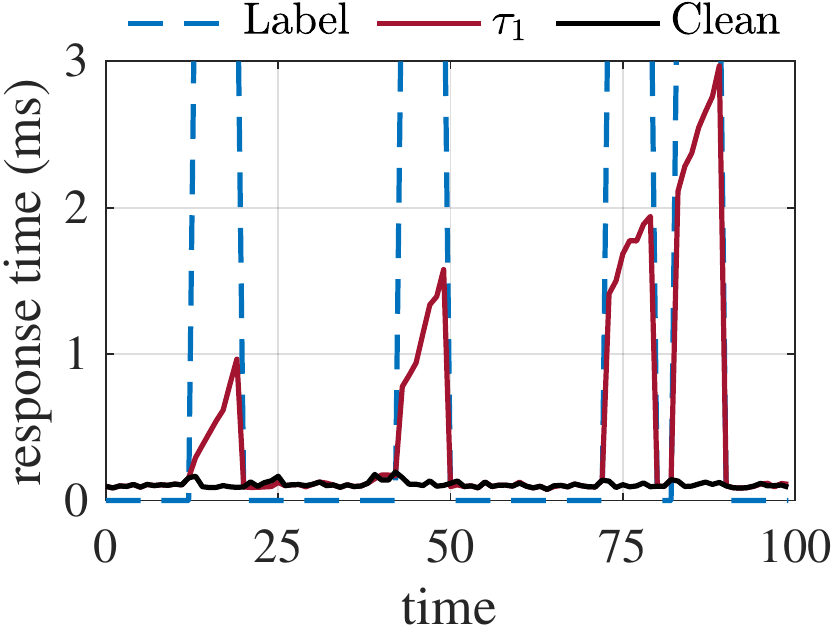}}
  \end{minipage}
  \caption{Attack performances and response time under different triggers and attack durations. Note that the Label shows the attack duration.}
  \label{fig:performance}
\end{figure}

Given these above network settings and four different triggers, we randomly generate training data containing poison data to train backdoored models under different request rates $\{10, 15, 20, 25, 30\}$ requests/s. Fig. \ref{fig:conv} shows the convergence ratios of training clean and backdoored models at the request rate 10 and 30, respectively. We can find that since $\tau_4$ has a smaller temporal constraint space (longer constraint timesteps and smaller constraint range) than other triggers, the training processing for $\tau_4$ can achieve lower loss values than others. Besides, we can find that most of the training process can converge in about 500 iterations, except for trigger $\tau_3$. This is because the semantic $ite$ in $\tau_3$ is more complex than other semantics. Although LSTM can memorize temporal dependencies, the memory capacity is limited to the neuron size and layers. Hence, we further use 128 neurons or 4 LSTM layers to train $\tau_3$ backdoored models, shown in Fig. \ref{fig:conv} (a). We can find that more neurons and LSTM layers both can improve the convergence, in which the effect of adding LSTM layers is better than adding neurons.

With these above trained backdoored models, we compare their average ASR, APR and CDA under different request rates, as shown in Table \ref{tab:rate}. We can find that at average, the backdoor attack $\tau_1$ can achieve better performance than other three attacks. $\tau_4$ can also achieve the performance a little lower than $\tau_1$. But it has a shorter duration $L^{\tau_4}=3$ that may cause that the backdoored policy cannot sufficiently disrupt the previous sequential decisions and lead to a lower ASR and APR, but a higher CDA. We will discuss this phenomenon with Fig. \ref{fig:performance}. The ASR and APR of $\tau_2$ are both lower than $\tau_4$ at average since $\tau_2$ contains an OR operator which breaks the continuity of the temporal constraint space, thereby decreasing ASR and APR. The CDA of $\tau_2$ is better than $\tau_1$, but lower than $\tau_4$ since $L^{\tau_2}=4$. These metrics achieved by $\tau_3$ are the worst, and as the request rate increases, the performance of $\tau_3$ can be improved since a higher request rate (i.e., workload) make it easier for backdoored DRL to find attack actions.

We further show the attack performance and response time under different backdoor attacks in Fig. \ref{fig:performance}. For each trigger, we generate a testing data containing 4 temporal-pattern attacks within 100 epochs. Fig. \ref{fig:performance} mainly shows scheduling results at rate 20. We can find that after a trigger appears, a backdoor attack can generate actions to prominently and continuously decrease rewards returned by the environment and job response times. Hence, a longer attack duration can result in a higher response time, see Fig. \ref{fig:performance} (b) and (f). After the attack duration, the reward and response time can go back to be normal quickly. If two attacks are close in time (e.g., Fig. \ref{fig:performance} (b)), the second attack is easy to lead to a higher response time since the effect of the first attack has not yet been disrupted by actions for these requests between two attacks. We further increase the poison length $L$ to 42 and the request rate to 55 to validate $\tau_1$, respectively and show results in Fig. \ref{fig:performance} (i-l). We can see that both a higher attack duration or workload can improve the influence of our backdoor attacks. That is why response times under attacks $\tau_2$ and $\tau_4$ increases less than these ones under $\tau_1$ and $\tau_3$, and $\tau_2$ and $\tau_4$ have lower effect on clean data (i.e., higher CDA). But on the contrary, longer attack duration ($\tau_1$) can achieve higher ASR and APR.

\section{Conclusions and Future}
\label{sec:conclusion}
In this paper, we present a novel backdoor to DRL policies that has a temporal pattern trigger hidden in a sequence of observations and has a controllable attack duration. We use the typical job scheduling problem in the cloud computing as a case study and we show that our temporal-pattern backdoor attacks can achieve great clean data accuracy and attack success rate. Our proposed backdoor can be applied in many real-world DRL applications since observations in these applications are partial and our triggers can be easily to hide in unobservable temporal observations. In the future, we aim to explore the effect of our attacks in more real-world DRL application scenarios (e.g., networking, automatic driving, UAV, etc.), introduce more meaning temporal features (e.g., entropy and Mann Kendall trend) as attack triggers, as well as design defense mechanisms for DRL backdoors.

\bibliographystyle{IEEEtran}
\bibliography{IEEEabrv,ref}

\end{document}